\documentclass[]{article}

\usepackage{geometry}
\usepackage{tikz}
\usepackage{cite}
\usetikzlibrary{calc}
\usetikzlibrary{arrows,shapes,chains}
\usepackage{xcolor}
\usepackage{amsfonts}
\usepackage{amsmath}
\usepackage{graphicx}
\usepackage{multirow}
\usepackage{algorithmic}
\usepackage{algorithm}
\usepackage{subfigure}
\usepackage{epstopdf}
\usepackage{url}
\usepackage{cleveref}
\crefname{equation}{}{}
\crefname{table}{TABLE}{TABLES}
\crefname{figure}{Fig.}{Figs.}
\crefname{section}{Section}{Sections}

\DeclareMathOperator*{\argmax}{arg\,max}

\newcommand{\cb}[1]{\ifmmode {\boldsymbol{#1}}\else ${\boldsymbol{#1}}$\fi}

\newcommand{\cp}[1]{\ifmmode {\mathcal{#1}}\else ${\mathcal{#1}}$\fi}

\newcommand{\red}{\color{black}}

\newcommand{\magentaso}{\color{black}}
\newcommand{\violetso}{\color{black}}
\newcommand{\orangesso}{\color{black}}

\title{Hyperspectral Image Classification with Deep Metric Learning and Conditional Random Field}
\author{Yi~Liang, Xin~Zhao, Alan J.X. Guo, and Fei Zhu
    \thanks{The work was supported in part by the National Natural Science Foundation of China under Grant 61701337 and the Natural Science Foundation of Tianjin under Grand 18JCQNJC01600. \emph{(Corresponding author: Alan J.X. Guo.)} 
    }
	\thanks{Y.~Liang, X.~Zhao, A.~Guo, and F.~Zhu are with the Center for Applied Mathematics, Tianjin University, China. (liangyi\_math;~zhaoxin\_zx;~jiaxiang.guo;~fei.zhu@tju.edu.cn) }
}

\begin{document}
	
\maketitle
\begin{abstract}
To improve the classification performance in the context of
hyperspectral image processing, many works have been developed based on two common strategies,
namely the spatial-spectral information integration and the utilization of neural networks.
However, both strategies typically require more training data than the classical algorithms, 
aggregating the shortage of labeled samples.
In this letter, we propose a novel framework that organically combines 
{\magentaso the} spectrum-based deep metric learning model
and the conditional random field algorithm.
The deep metric learning model is supervised by the center loss 
to produce spectrum-based features that gather more tightly {\magentaso in Euclidean space within classes}. 
{\magentaso The conditional random field with Gaussian edge potentials, which is firstly proposed for image 
segmentation tasks, is introduced to give the pixel-wise classification over the hyperspectral image 
by utilizing both the geographical distances between pixels and the Euclidean distances between the features 
produced by the deep metric learning model. 
}
The proposed framework is trained by spectral pixels at the deep metric learning stage
and utilizes the half handcrafted spatial features at the conditional random field stage.
This settlement alleviates the shortage of training data to some extent.
Experiments on two real hyperspectral images demonstrate the advantages of the proposed method in terms of both
classification accuracy and computation cost.
\end{abstract}
	
\section{Introduction}
Hyperspectral images (HSI) are usually acquired by spaceborne or airborne sensors,
recording the reflection {\magentaso spectra} or radiance spectra over hundreds of channels.
They are usually formatted as data cubes.
The height and width of an HSI data cube {\magentaso correspond} to the real world object under a specific resolution,
while the depth is decided by the channels of the sensors.
As a crucial task, the classification of HSI pixels attracts great attention for a long time \cite{lu2007survey,fauvel2013advances,li2018discriminant}.
Many early methods are based on classical machine learning algorithms and their variations,
for instance, principal component analysis (PCA)
\cite{prasad2008limitations, jiang2018superpca},
independent component analysis (ICA) \cite{villa2011hyperspectral},
linear discriminant analysis (LDA) \cite{bandos2009classification, li2011locality},
support vector machine (SVM)
\cite{melgani2004classification},
and sparse representation
\cite{chen2011hyperspectral, fang2014spectral}.

In recent years, neural networks (NN) have gained popularity in many applications related to machine learning, 
due to its power in {\magentaso generating} abstract representations from the original data. 
An increasing number of NN-based algorithms
have been adapted to HSI classification {\magentaso tasks} and achieved {\red impressive} results.
Representatives of the earlier models are stacked autoencoder (SAE)
\cite{chen2014deep}, and deep belief network (DBN) \cite{chen2015spectral}.
With the advances in deep learning, various deep models have been applied to HSI classification tasks, 
{\magentaso demonstrating their power in both processing spatial data and producing self-learned features.}
This category of algorithms mainly {\magentaso includes} convolutional neural network (CNN)
\cite{slavkovikj2015hyperspectral,chen2016deep, Jiao2017Deep},
recurrent neural network (RNN)
\cite{Mou2017Deep, liu2017bidirectional, zhang2018spatial}, and
deep metric learning (DML) \cite{guo2017spectral, guo2018cnn, cheng2018when}, to name a few.

The conditional random field (CRF) is a probabilistic graphical algorithm 
that enables to characterize the contextual information among the labels and the 
data~\cite{lafferty2001conditional}. As an important application of CRF, 
image segmentation has also attracted attention in
classifying HSI pixels~\cite{alam2016crf, pan2018high, niu2019deeplab, alam2018conditional}. 
In most of these works, {\magentaso the CRFs were integrated sequentially after the} CNNs as a post-processing step, processing the output features extracted by CNN encoders.
{\red
For example, in \cite{pan2018high}, a restricted CRF algorithm is applied to refine the 
superpixel classification {\magentaso from} a CNN to the final pixel-wise classification results.
In \cite{alam2018conditional}, the authors utilized {\magentaso a }CRF to improve the predictions on the CNN outputs and designed
a specific deconvolutional network to produce the final classifications.}


In this letter, a framework that combines the DML and CRF algorithms is proposed.
{\magentaso The DML model supervised with center loss is employed to extract spectrum-based 
	features from individual pixels. The CRF algorithm is applied to give final predictions 
	by modeling both the spatial and spectral information from the spectrum-based features 
	extracted by the DML model.}
To be more precise, our work has advantages in the following aspects:
\begin{itemize}
	\item  The intrinsic relations between DML and CRF help {\magentaso to improve }the classification accuracies.
	To the best of our knowledge, we are the first to introduce a framework that benefits from 
	the underlying connections between DML and CRF.
    \item The setting of employing a spectrum-based DML model and a handcraft spatial-based CRF algorithm keeps the framework simple.
	Compared to the CNN models, our framework is spectrum-based in the training phase and engages a simpler model structure, 
	thus alleviating the shortage of labeled HSI data raised in the CNN models \cite{chen2016deep,guo2017spectral}.
	\item {\magentaso In practice, the proposed framework shows high efficiency in computation cost, 
	for introducing the convolutional CRF (ConvCRF) \cite{teichmann2018convolutional}, 
    in which the CRF inferences are implemented on the GPU phase by convolutional operations.}
\end{itemize}

\section{Proposed Framework}\label{sec:ProposedModel}
As {\magentaso the }two main parts of the proposed framework, DML and CRF algorithms are firstly introduced separately. Substantially, an overview of the whole DML-CRF framework is presented.
\subsection{Deep Metric Learning} \label{subsec:dml}
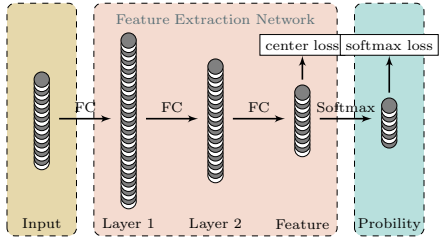
\begin{figure}
	{\linespread{1}
		\centering
		\scriptsize
		\tikzstyle{format}=[circle,draw,thin,fill=white]
		\tikzstyle{format_gray}=[circle,draw,thin,fill=gray]
		\tikzstyle{format_rect}=[rectangle,draw,thin,fill=white,align=center]
		\tikzstyle{arrowstyle} = [->,thick]
		\scalebox{0.77}
		{
			\begin{tikzpicture}[node distance=4mm,  auto,>=latex',  thin,  start chain=going below, every join/.style={norm}, scale=0.3]
			\definecolor{gray_so}{RGB}{88,110,117}
			\definecolor{yellow_so}{RGB}{181,137,0}
			\definecolor{cyan_so}{RGB}{42,161,152}
			\definecolor{orange_so}{RGB}{203,75,22}
			\filldraw[fill=yellow_so,rounded corners,fill opacity=0.33,style=dashed] (-2,-0.5) rectangle (2,12.9);
			\foreach \x/\xtext in {0, 2, 4, 6, 8, 10, 12, 14, 16, 18}
			{
				\node at (0,\x/4+3.75) [format] { };
				\node at (0,\x/4+0.25+3.75) [format_gray] { };
			}
			\filldraw[fill=orange_so,rounded corners,fill opacity=0.2,style=dashed] (3,-0.5) rectangle (17,12.9);
			\foreach \x/\xtext in {0,2,4,6,8,10,12,14,16,18,20,22,24,26,28,30,32,34,36}
			{
				\node at (5,\x/4+1.5) [format] { };
				\node at (5,\x/4+0.25+1.5) [format_gray] { };
			}
			\filldraw[fill=cyan_so,rounded corners,fill opacity=0.33,style=dashed] (18,-0.5) rectangle (22,12.9);
			\foreach \x/\xtext in {0,2,4,6,8,10,12,14,16,18,20,22,24}
			{
				\node at (10,\x/4+3) [format] { };
				\node at (10,\x/4+0.25+3) [format_gray] { };
			}
			\foreach \x/\xtext in {0, 2, 4, 6, 8, 10, 12}
			{
				\node at (15,\x/4+4.5) [format] { };
				\node at (15,\x/4+0.25+4.5) [format_gray] { };
			}

			\foreach \x/\xtext in {0, 2, 4, 6}
			{
				\node at (20,\x/4+5) [format_gray] { };
				\node at (20,\x/4+0.25+5) [format] { };
			}
			\node at (20,8/4+5) [format_gray] { };
			
			
			\draw[arrowstyle] (1,6.25) -- node {FC} (4,6.25);
			\draw[arrowstyle] (6,6.25) --  node {FC} (9,6.25);
			\draw[arrowstyle] (11,6.25) -- node {FC} (14,6.25);
			\draw[arrowstyle] (16,6.25) -- node {Softmax} (19,6.25);
			\node at (0,1.5) (n0) {};
			\node at (5,1.5) (n5) {};
			\node at (10,1.5) (n10) {};
			\node at (15,1.5) (n15) {};
			\node at (20,1.5) (n20) {};
			\node at (25,1.5) (n25) {};
			\node[below of =n0] (i) {Input};
			\node[below of =n5] (l1) {Layer 1};
			\node[below of =n10] (l2) {Layer 2};
			\node[below of =n15] (l3) {Feature};
			\node[below of =n20] (o) {Probility};
			\node at (15,10.5) [format_rect] (centerloss) {center loss};
			\node at (20,10.5) [format_rect] (loss) {softmax loss};
			\draw[arrowstyle] (15,8.5) to (centerloss.south);
			\draw[arrowstyle] (20,7.75) to (loss.south);
			\node at (10,12) [rounded corners,style=dashed,color=gray_so] (input_section) {Feature Extraction Network};
			\end{tikzpicture}
		}
		\caption{\label{mlp_struct} Structure of the spectral feature {\red extraction} network DML~\cite{guo2017spectral}.}
	}
\end{figure}

In \cite{guo2017spectral}, the center loss proposed in the 
deep metric learning model \cite{wen2016discriminative} 
was first introduced to the HSI classification tasks. A 3-layered fully connected network was built 
to extract spectral features from the input data. {\magentaso As illustrated in \cref{mlp_struct}, 
the model is jointly supervised by cross-entropy loss (also called softmax loss) and center loss. 
Under this settlement, the extracted features from the same class gather more tightly in Euclidean space.}
This model is adopted to encode the spectrum in our work.

Throughout this letter, we use $(\mathbf{x}_i,y_i)$ to denote 
the pixel $\mathbf{x}_i$ with the label $y_i$, from the HSI $\mathbf{X}$.
Let $f(\cdot)$ be the function defined by the neural network, {\magentaso whose values are} the extracted features.
Use $\tilde{p}(y|\mathbf{x}_i)$ to denote the {\magentaso predicted} probability distribution 
{\magentaso that is calculated} by applying the softmax function
on the extracted features $f(\mathbf{x}_i)$.
During the training stage, a joint loss $\mathcal{L}$ that sums the center loss $\mathcal{L}_c$
and the cross-entropy loss $\mathcal{L}_s$ is engaged.
{\red As the key part of DML}, the center loss $\mathcal{L}_c$ is defined to measure the Euclidean distance between the
produced features $f(\mathbf{x}_i)$ and its class centers
$\mathbf{c}_{y_i}$, as
\begin{equation} \label{euclidean_dml}
	\mathcal{L}_c = \sum_i || f(\mathbf{x}_i) - \mathbf{c}_{y_i} ||_2,
\end{equation}
where the class centers are formulated as
\[
 \mathbf{c}_{y_i} = \mathrm{average}(\{f(\mathbf{x}_k)|\,y_k = y_i\}).
\]

At the testing stage, samples $\mathbf{x}_i$ are fed to the neural network $f(\cdot)$.
The outputs{, \magentaso which} include both the extracted
feature $f(\mathbf{X})$ and the predicted probability distribution $\tilde{p}(y|\mathbf{x}_i)$, 
{\magentaso are} collected for the subsequential CRF step.

\subsection{Conditional Random Field}

The CRF algorithm plays an important role in image segmentation, with the merit of
exploiting the global context information
\cite{lafferty2001conditional, krahenbuhl2011efficient, teichmann2018convolutional}.
In this letter, we use the CRF with Gaussian edge potentials to fuse the spatial-spectral information, 
and give reasonable {\magentaso pixel-wise} predictions of the HSI.
The notations in this letter mainly follow the models of fully connected CRF in \cite{krahenbuhl2011efficient} and ConvCRF in \cite{teichmann2018convolutional}.

Let $\hat{y}$ be a choice of predictions over all the pixels in an HSI, the probability of $\hat{y}$ is calculated from
an energy function $E(\hat{y}|f(\mathbf{X}))$ by a Gibbs distribution:
\[
P(\hat{y}|f(\mathbf{X})) = \frac{\mathrm{exp}(-E(\hat{y}|f(\mathbf{X})))}{Z(f(\mathbf{X}))},
\]
where $Z(f(\mathbf{X}))$ is the partition function \cite{lafferty2001conditional}.
In this algorithm, the energy function $E(\hat{y}|f(\mathbf{X}))$ is set to have two parts, with
\[
E(\hat{y}|f(\mathbf{X})) = \sum_{i} \psi_u(\hat{y}_i|\mathbf{x}_i) + \sum_{i<j}\psi_p(\hat{y}_i,\hat{y}_j|\mathbf{x}_i,\mathbf{x}_j),
\]
where $\psi_u(\hat{y}_i|\mathbf{x}_i)$ is the unary potential and 
$\psi_p(\hat{y}_i,\hat{y}_j|\mathbf{x}_i,\mathbf{x}_j)$
is the pairwise potential.
As in most applications of CRF, the unary potential is set {\magentaso to be} the cost of a pixel $\mathbf{x}_i$ taking label $\hat{y}_i$,
which is
\begin{equation}\label{unary}
\psi_u(\hat{y}_i|\mathbf{x}_i) = -\mathrm{log}(\tilde{p}(\hat{y}_i|\mathbf{x}_i)).
\end{equation}
The pairwise potential is set to be
	\begin{equation}\label{GaussinEdgeP}
	\psi_p(\hat{y}_i,\hat{y}_j|\mathbf{x}_i,\mathbf{x}_j) = \mu(\hat{y}_i,\hat{y}_j)(w^{\mathrm{app}}k_{\mathrm{app}}+
	w^{\mathrm{smo}}k_{\mathrm{smo}}),
	\end{equation}
	{\red where $\mu(\hat{y}_i,\hat{y}_j)$ is called compatibility function and given by the Potts model
	$\mu(\hat{y}_i,\hat{y}_j)=|\hat{y}_i\neq \hat{y}_j|$, 
	the $k_{\mathrm{app}}$ and $k_{\mathrm{smo}}$ are respectively termed the appearance and smooth kernels, 
	and the $w^{\mathrm{app}}$ and $w^{\mathrm{smo}}$ are linear combination weights.}
	If we denote the position of $\mathbf{x}_i$ as $p_i$, the appearance kernel $k_{\mathrm{app}}$ in \eqref{GaussinEdgeP} is defined as
	\begin{eqnarray}
	&&k_{\mathrm{app}}(\hat{y}_i,\hat{y}_j|\mathbf{x}_i,\mathbf{x}_j) \nonumber \\
	&=&\mathrm{exp}\left(-\frac{|p_i-p_j|^2}{2\theta_\alpha^2}-\frac{|f(\mathbf{x}_i)-f(\mathbf{x}_j)|^2}{2\theta_\beta^2}\right) \label{euclidean_crf}
	\end{eqnarray}
	and the smoothness kernel $k_{\mathrm{smo}}$ is
	\[k_{\mathrm{smo}}(\hat{y}_i,\hat{y}_j|\mathbf{x}_i,\mathbf{x}_j) =
	\mathrm{exp}\left(-\frac{|p_i-p_j|^2}{2\theta_\gamma^2}\right).
	\]
As stated in~\cite{krahenbuhl2011efficient}, the appearance kernel is based on the observation 
that neighboring pixels with similar features tend to be from the same class, 
while the smoothness kernel helps to eliminate small isolated regions.
It is noteworthy that the pairwise potential integrates both the spectral information and 
geographical information.

Mathematically, the final prediction is obtained by
\begin{equation*}
	y^* = \argmax_{\hat{y}}P(\hat{y}|f(\mathbf{X})),
\end{equation*}
which is, however, hard to compute.
Usually, a method of mean field approximation \cite{krahenbuhl2011efficient} is used to 
approximately calculate the results.
In \cite{teichmann2018convolutional}, the authors assumed that the pairwise potentials
only take effect when the Manhattan distance between $\mathbf{x}_i$ and $\mathbf{x}_j$ is less than
the so-called filter-size $k$. Under this assumption, the mean field inference algorithm  
{\magentaso could be} implemented on the GPU phase and
calculated more efficiently. This inference algorithm is called
ConvCRF. Readers may refer \cite{krahenbuhl2011efficient, teichmann2018convolutional}
for more detailed definitions and calculations of CRF.

\subsection{\red A Summary}
{\red In general, we first use a DML model to generate spectrum-based features $f(\mathbf{x}_i)$, as well as
the preliminary predictions $\tilde{p}(y|\mathbf{x}_i)$. 
Then, {\magentaso the preliminary predictions $\tilde{p}(y|\mathbf{x}_i)$ are reformulated as the unary
potentials of CRF by~\eqref{unary}.} The pairwise potentials, which include the appearance and smooth kernels, 
are expressed by~\eqref{GaussinEdgeP} using features $f(\mathbf{x}_i)$ and 
{\red the corresponding pixels' positions $p_i$}.
Finally, the ConvCRF algorithm is adopted to make CRF inference, 
producing the final predictions {\magentaso over all the pixels in an HSI}.}

{\red In essence, it is the intrinsic connection between the center loss of DML and the appearance kernel {\magentaso of} CRF that contributes
to the performance of the proposed framework. Compared to the features extracted by the conventional NN models, 
the features extracted by DML with center loss gather more tightly
{\magentaso in Euclidean space within the same class}, {\em i.e.}, 
pixels from the same class tend to be encoded as more similar features.
Meanwhile, the appearance kernel \eqref{euclidean_crf} is designed to rely on 
the Euclidean distances between features $f(\mathbf{x}_i)$.}
When compared to CRFs that rely on raw pixel spectra or features from plain NN models,
the existence of center loss in DML {\magentaso rationalizes} the CRF algorithm in our 
framework and enhance the final classification results.

\section{Experiments}\label{sec: Experiments}
\subsection{Datasets Description}
The experiments are carried out on two well-known HSI datasets,
namely the Pavia University scene and the Salinas scene collected by the
ROSIS sensor and the AVRIS sensor, respectively{\footnote{The datasets are available online: \url{http://www.ehu.eus/ccwintco/index.php?title=Hyperspectral_Remote_Sensing_Scenes}}}.
The Pavia University scene used in experiments has a size
$(610, 340, 103)$ in $(\mathrm{Height},\mathrm{Width},\mathrm{Bands})$.
The spatial resolution is $1.3\,\mathrm{m}$, 
while the band depth covers the wavelength from $0.43\,\mathrm{\mu m}$ 
to $0.86\,\mathrm{\mu m}$ with $12$ noisy and water absorption bands removed.
{\red Regarding the Salinas scene, the size of the image is $(512, 217, 204)$ 
after removing $20$ water absorption bands. 
The spatial resolution is $3.7\,\mathrm{m}$, and the spectra cover a bandwidth 
range from $0.4\,\mathrm{\mu m}$ to $2.5\,\mathrm{\mu m}$.}

\subsection{Experimental Settings and Results}
To show the advantage of combining DML and CRF, {\red contrast} experiments are carried out by 
{\magentaso implementing} DML, NN-CRF, and the proposed framework DML-CRF.
Here, the only difference between the NN model and the DML model is the absence of 
center loss in the former. Moreover, two state-of-art methods,
namely $3$D-CNN \cite{chen2016deep} and CSFF (DML-CSFF) \cite{guo2018cnn} are also compared 
as baselines of HSI classification algorithms.
Experiments are {\magentaso performed} with deep learning platforms Caffe \cite{jia2014caffe} and PyTorch \cite{paszke2017automatic},
on a machine equipped with CPU of Intel Xeon E5-2660@2.6GHz and
GPU of NVIDIA TitanX.

As for the DML model in DML and DML-CRF, the length of the extracted feature is set to be 
$\mathrm{len}(f(\mathbf{x_i})) = 32$. 
The hyperparameters, such as learning rate, balance weight $\lambda$, and {\em etc.}, 
are all chosen as {\magentaso their default values in the original paper \cite{guo2017spectral}}.
Regarding the CRF algorithm in NN-CRF and DML-CRF, there are five parameters 
$w^{\mathrm{app}}$, $w^{\mathrm{smo}}$,
$\theta_\alpha$, $\theta_\beta$, and $\theta_\gamma$.
According to~\cite{krahenbuhl2011efficient},
the performances of CRF in terms of
classificationy are {\red relatively} robust to these five parameters. Therefore, the default setting,
\[
w^{\mathrm{app}} = 10,\, w^{\mathrm{smo}} = 3,\, \theta_\alpha = 0.1,\, \theta_\beta = 80,\, \theta_\gamma = 3,
\]
in \cite{krahenbuhl2011efficient,teichmann2018convolutional} are used directly.
The only hyperparameter {\magentaso that needs} to be set is the filter-size $k$ in ConvCRF, which is chosen as $k = 7$ for Pavia University scene and
$k=15$ for Salinas scene.
{\red An analysis of these variables is given in \cref{p_a}.}
The comparing methods $3$D-CNN and CSFF (DML-CSFF) are implemented by following 
{\magentaso their original papers }~\cite{chen2016deep,guo2018cnn}.

If not otherwise specified, the training samples used in all the experiments 
follow the same preprocessing procedure.
Each dataset is firstly normalized to have zero mean and unit variance. 
The training set is formed by randomly chosen $200$ pixels per class.
For $3$D-CNN, $200$ HSI patches from each class are randomly chosen instead.
To avoid overfitting effects, virtual samples are generated {\magentaso by the linear combinations of the pixels from the same class, 
with formula}
$\tilde{\mathbf{x}} = q\mathbf{x}_1 + (1-q)\mathbf{x}_2.$
{\magentaso They are adopted in the training stages of all the aforementioned neural networks.}
The classification {\magentaso performances are} evaluated by three metrics, namely overall accuracy (OA), 
average accuracy (AA), and the kappa coefficient ($\kappa$).
{\orangesso Briefly, the metric OA is the percentage of correctly classified samples over all the 
testing samples, 
the metric AA is calculated by averaging the classification accuracies from each class, 
and the coefficient $\kappa$ 
measures the agreement between the predicted labels and groundtruth labels by the formula
\[
	\kappa = \frac{p_o-p_e}{1-p_e}.	
\]
In this formula, the notation $p_o$ represents the chance that the predicted label agrees with groundtruth label, 
which is the overall accuracy (OA), while $p_e$ is the hypothetical probability of chance agreement. 
Assume we have the predicted distribution which has chance $p_p(i)$ to output a predicted label $i$, 
and the groundtruth distribution 
which has chance $p_g(i)$ to output a groundtruth label $i$, $p_e$ is then calculated by 
\[
	p_e = \sum_{i} p_p(i)p_g(i).	
\]}

The classification results with mean and standard deviation over five runs {\red are} reported in \cref{result_table}. 
As shown in the first three columns, the absence of either DML or CRF deteriorates the classification accuracies.
Compared to the state-of-the-art methods, the proposed DML-CRF still leads to comparable results.
{\red The proposed DML-CRF outperforms $3$D-CNN with a large margin on both datasets.}
When compared to CSFF, DML-CRF performs better in all the metrics
on the Pavia University scene.
{\red
On the Salinas scene, DML-CRF {\red surpasses} CSFF in terms of AA, but is slightly inferior to CSFF in terms of OA and $\kappa$.}
The testing {\magentaso times of the comparing methods are} given in \cref{time_table}.
We observe the DML-CRF is overwhelmingly faster than CSFF and several times faster than  $3$D-CNN, thanks to the implementation of ConvCRF on
GPU. 

{\violetso 
In the proposed DML-CRF framework, the parameters in the DML model are trained by spectral data, 
while the parameters in the CRF algorithm are set directly. 
Compared to the most of the spatial-spectral algorithms which use the HSI patches as training data, 
only the spectral data is engaged in the training of DML-CRF. This alleviates the shortage of 
HSI data in one sense. Also, the algorithm DML-CRF engages a simple and spectrum-based DML model, 
hence it has fewer parameters than the spatial-spectral algorithms which usually use 
multiple CNN layers as the model structure. 
Typically, a model with less trainable parameters 
tends to have less overfitting issues, therefore it performs better with insufficient training data. 
In this letter, the training datasets of DML-CRF and $3$D-CNN are set to have the same 
cardinalities. 
Comparison between the classification accuracies of DML-CRF and $3$D-CNN in \cref{result_table} 
partially confirms our hypothesis mentioned above.
}

\begin{table*}
	\centering
	\caption{\label{result_table}Classification accuracies (averaged over 5 runs)  of DML, NN-CRF,
		DML-CRF, $3$D-CNN, and CSFF on Pavia University scene and Salinas scene}
	\begin{tabular}{c|c|c|c|c||c|c}
		\hline \hline
		\multicolumn{2}{c|}{}	& DML 	& NN-CRF 	& DML-CRF 	& $3$D-CNN 	& CSFF \\ \hline
		\multirow{3}{*}{Pavia Univ.} 	& OA($\%$)       & $93.67\pm0.31$ & $98.78\pm0.15$ & $99.10\pm0.10$ & $98.14\pm0.10$   &$98.90\pm0.14
		$      \\ \cline{3-7}
										& AA($\%$)       & $93.64\pm0.22$ & $97.53\pm0.18$ &$98.72\pm0.20$ &$97.32\pm0.68$   &$98.49\pm0.13
										$      \\ \cline{3-7}
										& $\kappa$		& $0.9153\pm0.0041$& $98.38\pm0.20$ & $0.9880\pm0.0014$&$0.9687\pm0.0015$&$0.9852\pm0.0018
										$      \\ \hline
		\multirow{3}{*}{Salinas}     	& OA($\%$)       & $92.72\pm0.44$ & $97.86\pm0.28$ &$98.12\pm0.21$ &$95.91\pm0.87$   &$98.53\pm0.34$      \\ \cline{3-7}
										& AA($\%$)       & $97.09\pm0.16$ & $99.25\pm0.11$ &$99.26\pm0.08$ &$98.79\pm0.29$   &$99.02\pm0.20$      \\ \cline{3-7}
										& $\kappa$ 		& $0.9186\pm0.0049$&$97.62\pm0.31$        &$0.9791\pm0.0024$         &$0.9480\pm0.0108$&$0.9835\pm0.0038$      \\ \hline \hline
	\end{tabular}
\end{table*}

\begin{table}
	\centering
	\caption{Comparison of testing time (in seconds)}\label{time_table}
	\begin{tabular}{c|c|c|c}
		\hline \hline
		{}            &  3D-CNN  &  CSFF& DML-CRF\\
		\hline
		Pavia Univ.  & $50.94$ & $1779.45$&$8.68$\\
		\hline
		Salinas &$45.69$ &  $5453.36$& $17.51$\\
		\hline \hline
	\end{tabular}
\end{table}

\begin{figure*}
	\centering
	\graphicspath{{Figures/}}
		\includegraphics[trim = 0mm 3mm 5mm 2mm, clip,width=0.32\textwidth]  {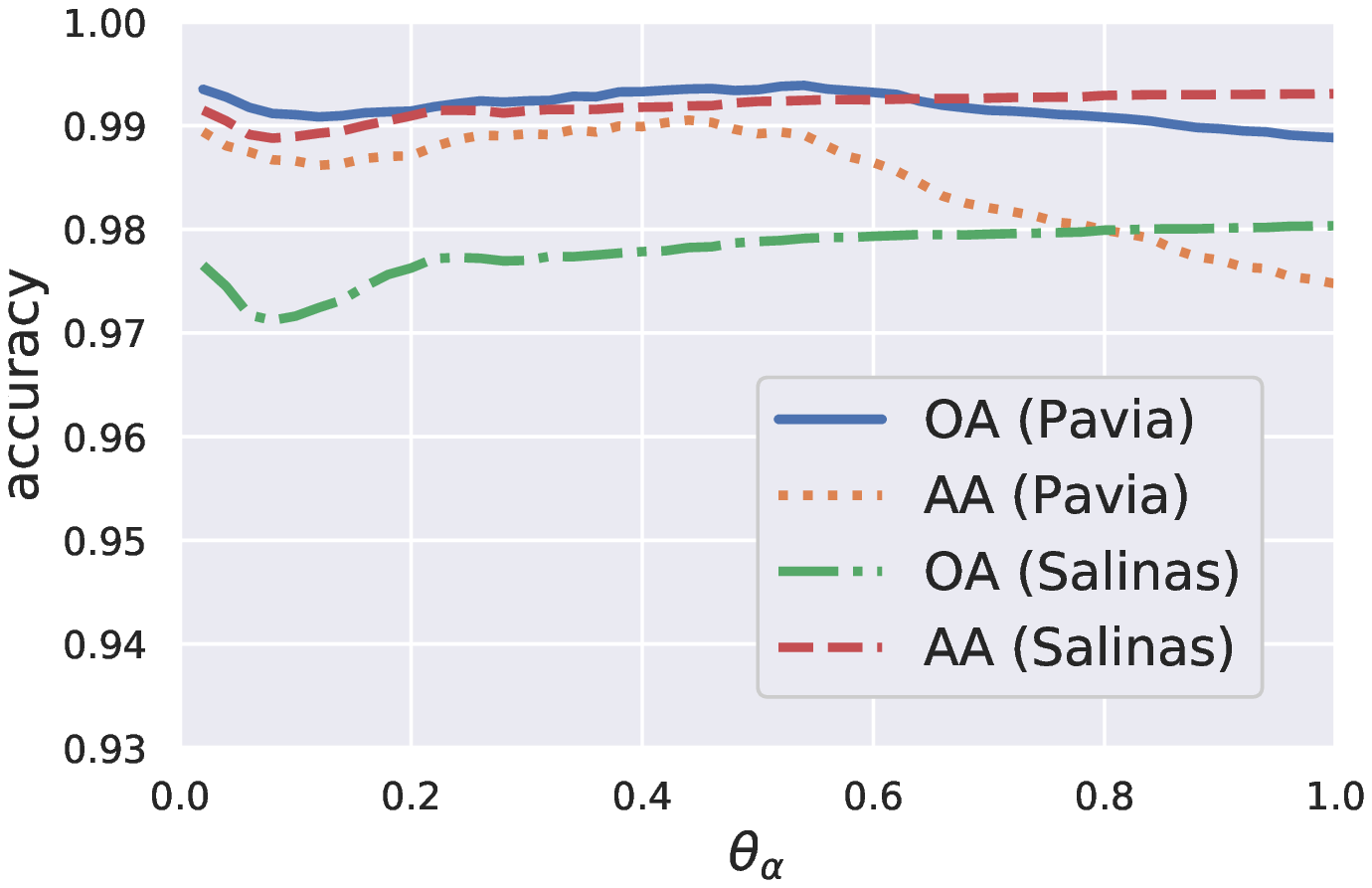}
	    \includegraphics[trim = 0mm 3mm 5mm 2mm, clip,width=0.32\textwidth]  {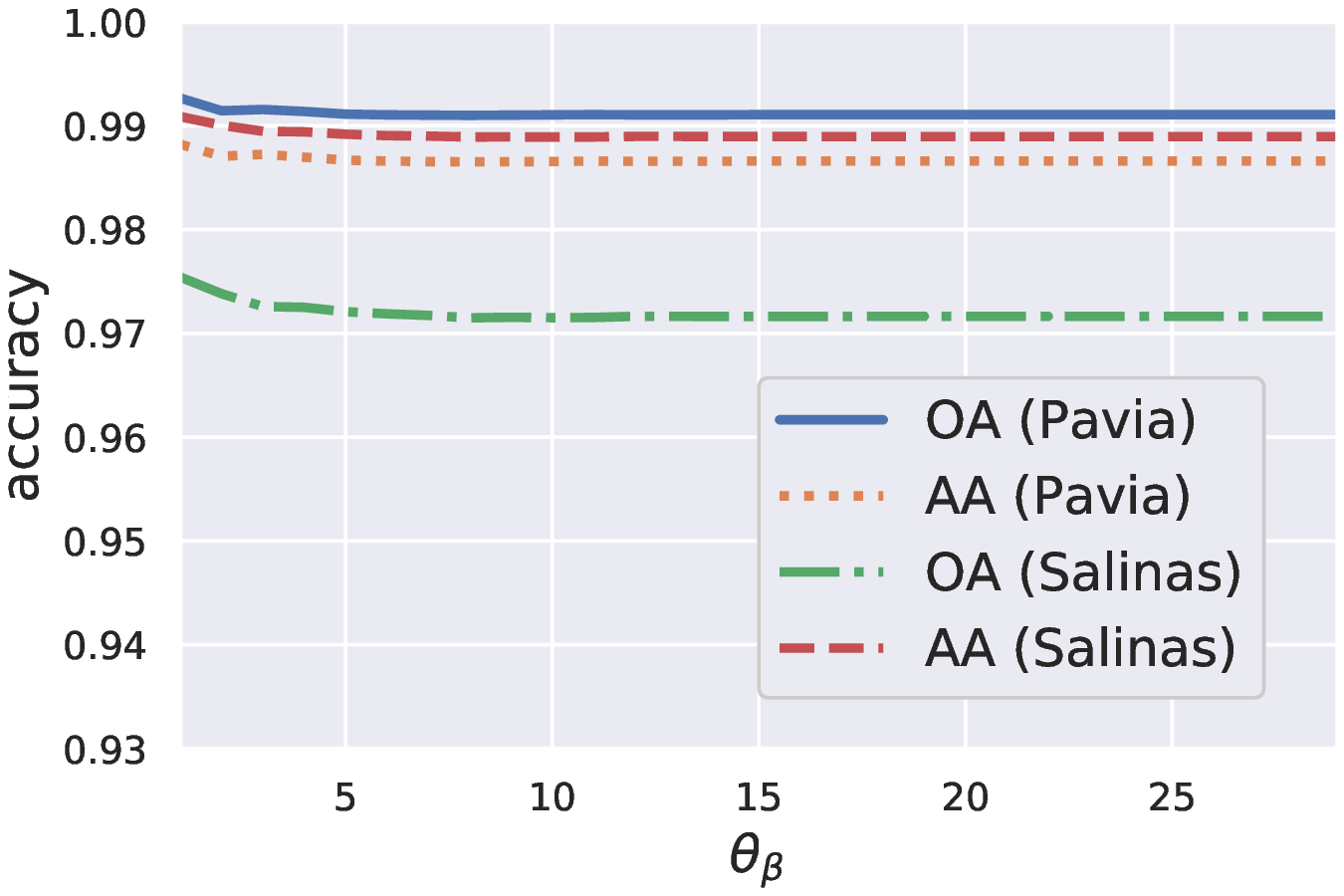}
        \includegraphics[trim = 0mm 3mm 5mm 2mm, clip,width=0.32\textwidth]  {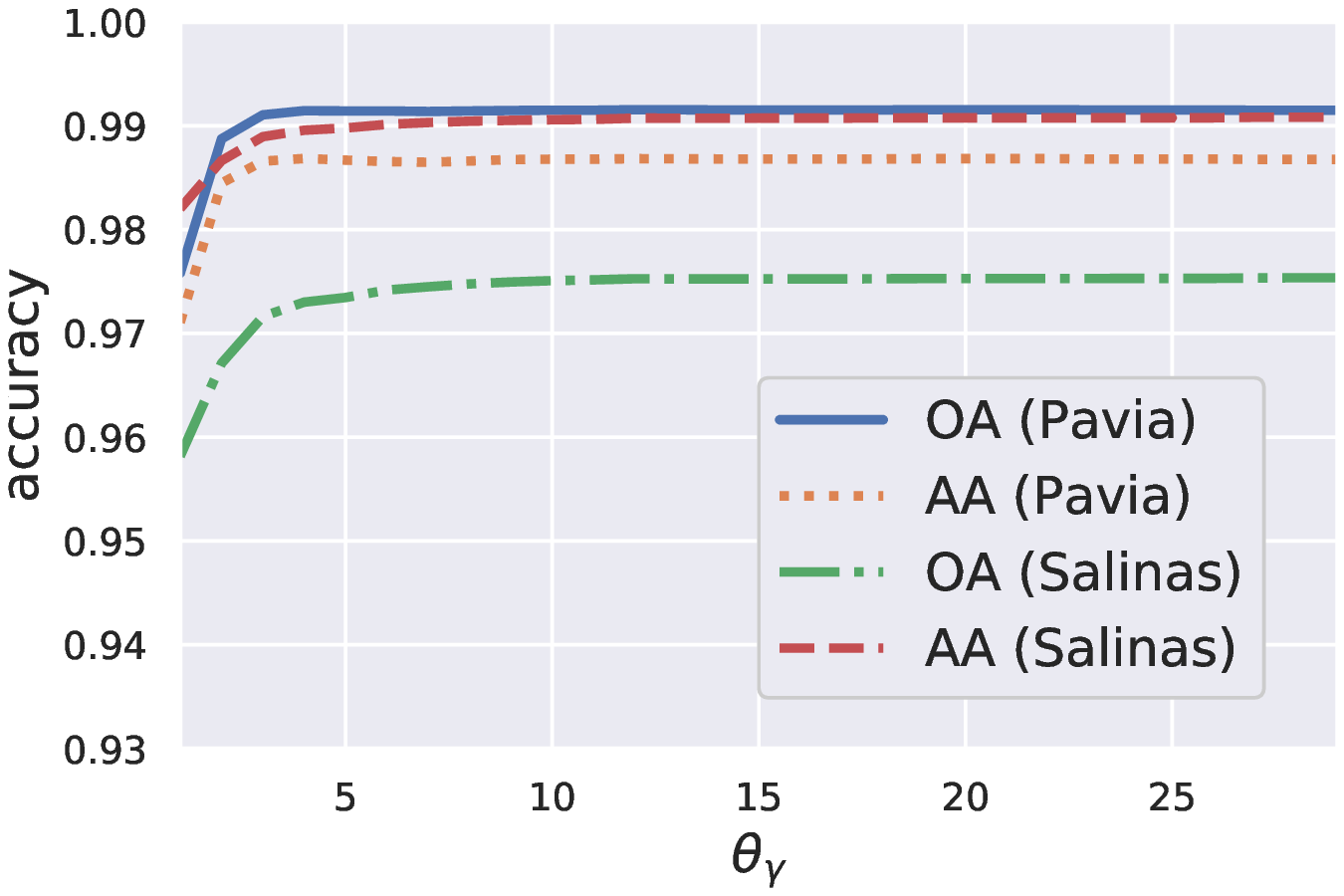}
        \includegraphics[trim = 0mm 3mm 5mm 2mm, clip,width=0.32\textwidth]  {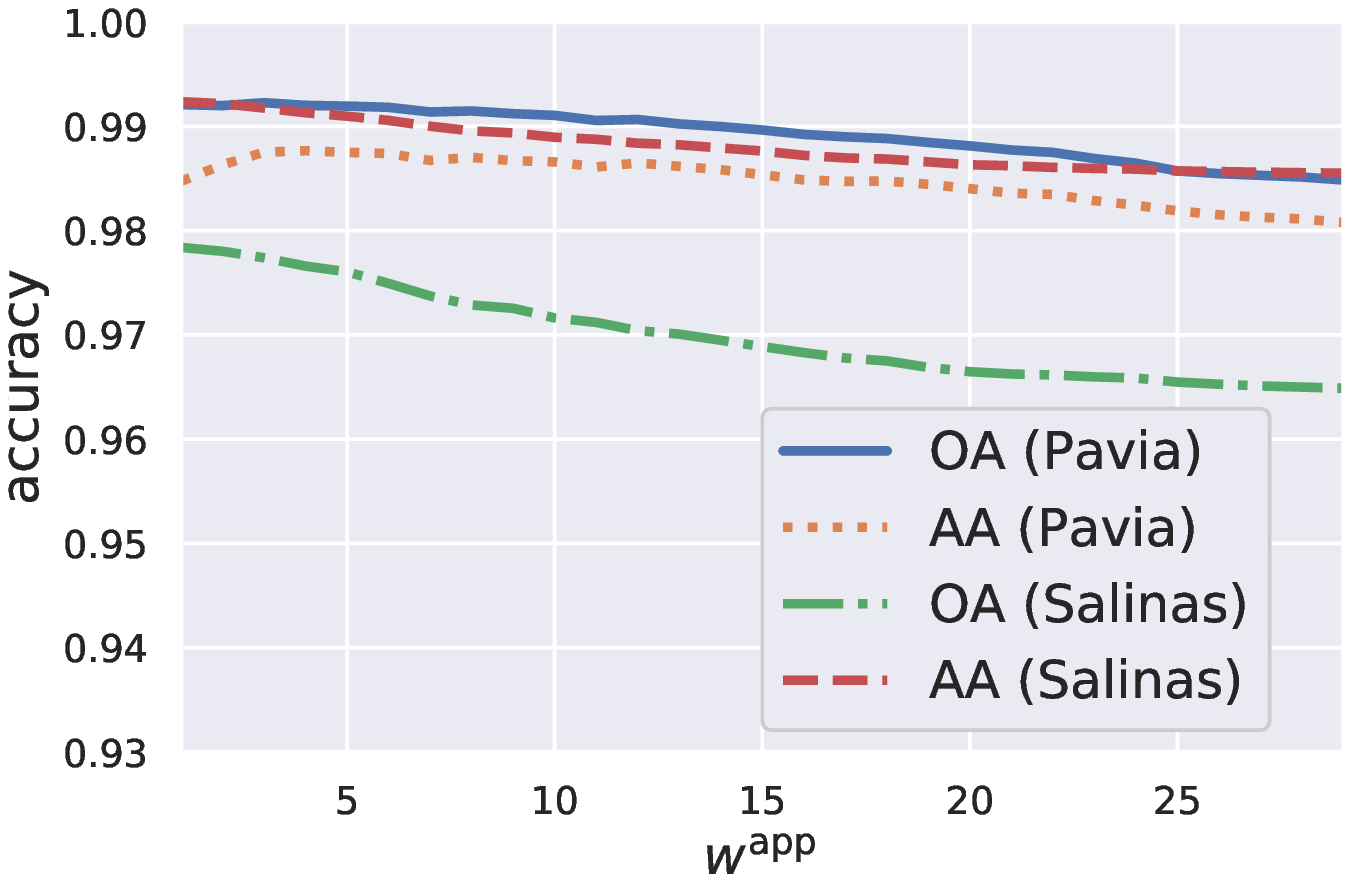}
        \includegraphics[trim = 0mm 3mm 5mm 2mm, clip,width=0.32\textwidth]  {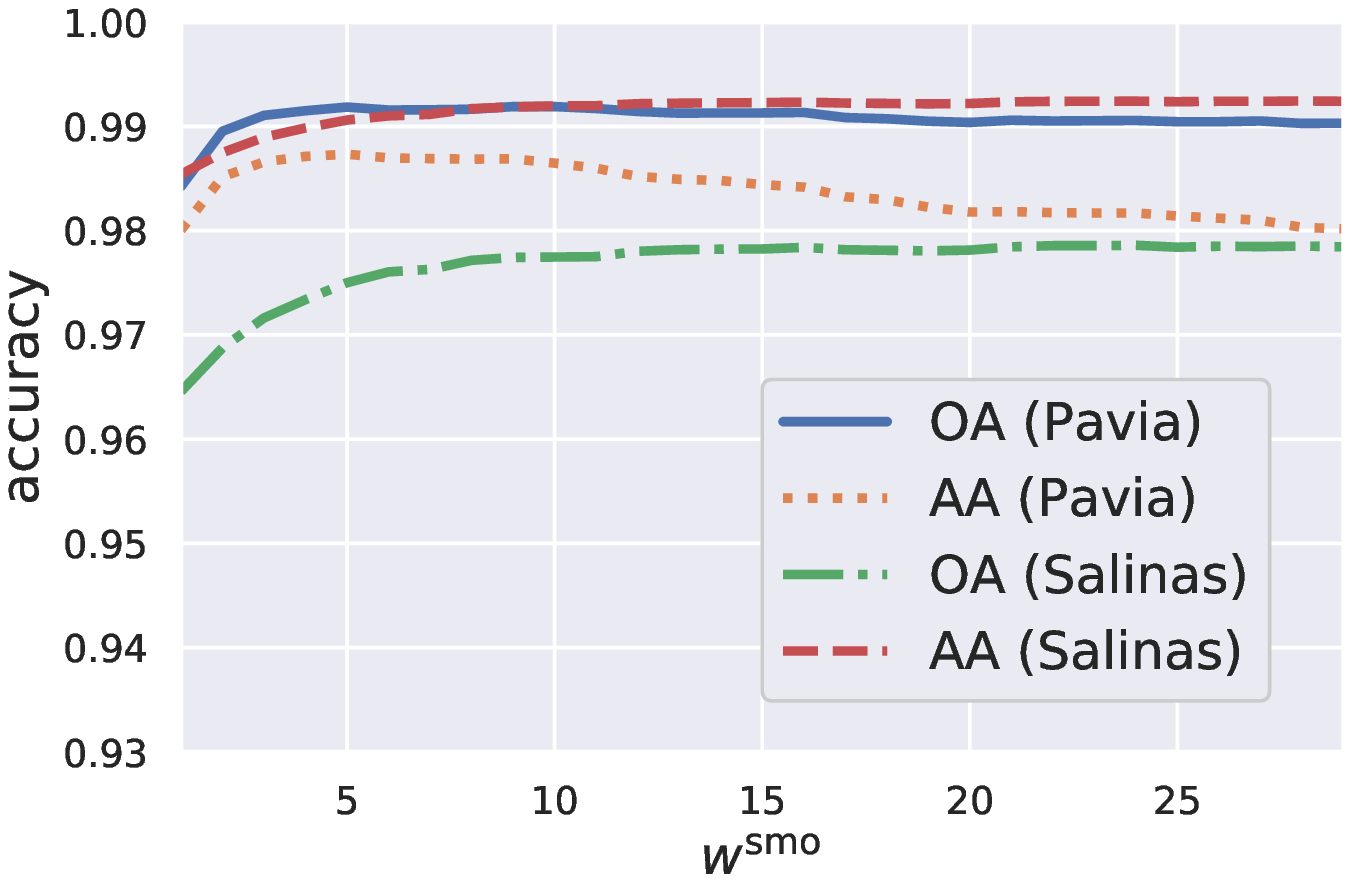}
	\caption{\label{v_p} Classification accuracies in terms of AA and OA, along with varying parameters $w^{\mathrm{app}}$, $w^{\mathrm{smo}}$, $\theta_\alpha$, $\theta_\beta$, and $\theta_\gamma$, on Pavia University and Salinas scenes. }
\end{figure*}

\begin{figure}
	\centering
	\graphicspath{{Figures/}}
		\includegraphics[trim = 0mm 3mm 5mm 3mm, clip,width=0.37\textwidth]  {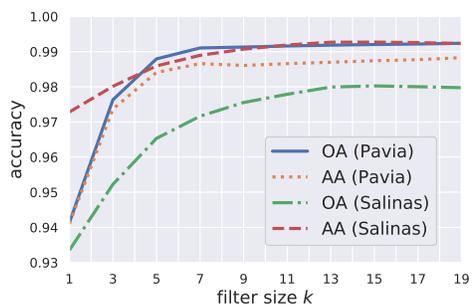}
	\caption{\label{v_hp} Classification accuracies in terms of AA and OA, along with varying hyperparameter $k$, on Pavia University and Salinas scenes. }
\end{figure}

\subsection{Parameter Optimization}\label{p_a}
This subsection mainly discusses the {\magentaso effects of different choices of 
parameters and hyperparameters in CRF.} 
For {\magentaso the hyperparameters in the DML model, details on their behaviors of them 
can be found in \cite{guo2017spectral}.}

{\magentaso To verify the robustness of CRF to the parameters
$w^{\mathrm{app}}$, $w^{\mathrm{smo}}$, $\theta_\alpha$, $\theta_\beta$, and $\theta_\gamma$, 
we anchor the default values by
$w^{\mathrm{app}} = 10,\, w^{\mathrm{smo}} = 3,\, \theta_\alpha = 0.1,\, \theta_\beta = 80,\, \theta_\gamma = 3$.
Under this setting, we perform several experiments by varying every single parameter at one time.}
The relationships between the {\magentaso parameters and the classification performances} are presented in \cref{v_p}.
It is obvious that the classification {\magentaso performances are} relatively robust to the parameters.

Regarding the only hyperparameter $k$, {\magentaso which is the filter-size} in ConvCRF, it controls the size of the spatial information that CRF takes into account.
The effect of $k$ on the classification accuracies are shown in \cref{v_hp}.
As expected, larger filter-sizes lead to higher accuracies, but also require more cost of computation.

\section{Conclusion}\label{sec: Conclusion}
In this letter, we proposed a framework that combines DML and CRF. 
The DML model is used to extract features from pixels of HSIs. The advantage of center loss
reduces the Euclidean distances between the extracted features which share the same class label.
Later, the CRF algorithm is applied to give predictions over the whole HSI by using both the extracted features
and {\red their} position information.
Contrast experiments demonstrated that the absence of either DML or CRF declines the classification performances.
Moreover, the proposed framework {\red provides comparable results to the state-of-art methods in both 
classification accuracies and computation cost.}
Additional experiments are performed to show the effects of varying parameters and hyperparameters 
on the classification accuracies.


\bibliographystyle{IEEEtran}
\bibliography{AlanGuo,bib_fei,crf}

\end{document}